\documentclass{article}
\usepackage[numbers]{natbib}
\usepackage{graphicx}

\usepackage[T1]{fontenc}
\usepackage{times} 
\usepackage{mathptmx} 
\usepackage[a4paper,margin=1in]{geometry} 

\usepackage{natbib} 
\usepackage{fancyhdr}
\usepackage[font=small,labelfont=bf]{caption} 

\usepackage{algorithm}
\usepackage{algpseudocode}
\usepackage{authblk}

\usepackage{amsmath}
\usepackage{amssymb}

\usepackage{graphicx} 
\usepackage{subcaption} 

\begin{document}

\title{Probabilistic Joint Recovery Method for CO$_2$ Plume Monitoring}

\author{
Zijun Deng, Rafael Orozco, Abhinav Prakash Gahlot, Felix J. Herrmann \\
    \small{\textit{Georgia Institute of Technology, GA}}
}

\date{}
\maketitle

\section{Introduction}
In order to curve current trends in climate change, it is crucial to reduce atmospheric CO$_2$ emissions \citep{ringrose2023storage}. Carbon Capture and Storage (CCS) is recognized as one of the only scalable net-negative CO$_2$ technologies. During injection, accurate prediction of fluid flow patterns in CCS is a challenging task, particularly due to uncertainties in CO$_2$ plume dynamics and reservoir properties. Monitoring techniques such as seismic imaging become imperative to understand the evolution of the CO$_2$ plume. Although previous time-lapse imaging methods such as the Joint Recovery Method (JRM) \citep{oghenekohwo2015CSEGctl, yin2021compressive, gahlot2023IMAGEtsm} have provided valuable information, however they do not communicate uncertainty thus are limited as tools for decision making. To address this, we propose the Probabilistic Joint Recovery Method (pJRM) that computes the posterior distribution at each monitoring survey while leveraging the shared structure among surveys through a shared generative model \citep{heckel2019deepdecoderconciseimage}. By computing posterior distributions for surveys, this method aims to provide valuable uncertainty information to decision makers in CCS projects, augmenting their workflow with principled risk minimization.

\subsection{Previous Work}
The JRM framework improves time-lapse imaging by leveraging the assumption that time-lapse seismic data shares common features between different surveys \citep{oghenekohwo2015CSEGctl,yin2021compressive} thus the joint inversion of these common features will produce a better result than independent inversion. This assumption is true for CO$_2$ monitoring due to the fact that large regions of the reservoirs remain unchanged i.e. regions where there is no CO$_2$ plume. However, the ill-posed nature of wave-based imaging used during time-lapse imaging (e.g. RTM, LS-RTM, FWI) introduces significant uncertainties that deterministic methods fail to account for. This work addresses this challenge by explicitly modeling these uncertainties, enabling more reliable subsurface property estimation.

\section{Method}
Inspired by previous work on reconstructing black holes utilizing shared features \citep{leong2023ill}, we adopted a similar framework to solve $N$ time-lapse problems jointly. Our approach takes $N$ seismic measurements as input and produces $N$ posterior distributions of the reconstructed plumes as output. To play the probabilistic counterpart of the common component in the JRM framework, the pJRM uses a Shared Generative Model (SGM) that extracts common features from the measurements. The innovation components (i.e., features unique to each survey) are captured by providing distinct latent distributions to the SGM for each monitoring survey.

\subsection{Probabilistic Joint Recovery Model}
\begin{figure}[!htb]
\centering
\begin{subfigure}{0.48\textwidth}
    \centering
    \includegraphics[width=\textwidth]{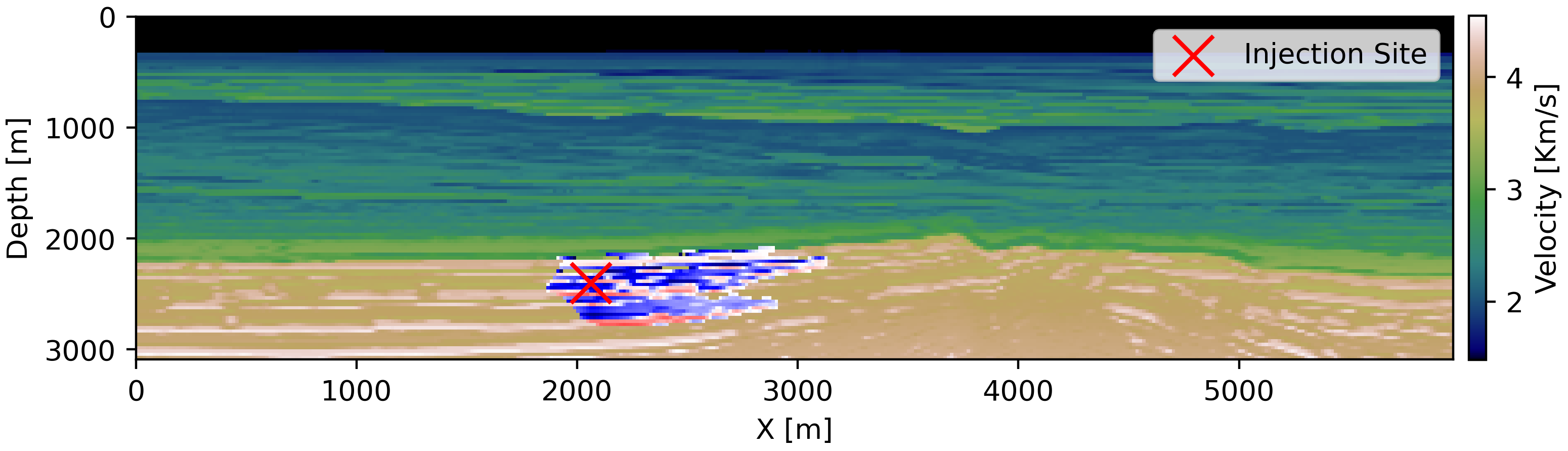}
    \caption{Synthetic earth model and unknown plume}
    \label{fig:earth}
\end{subfigure}
\\
\begin{subfigure}{0.48\textwidth}
    \centering
    \includegraphics[width=\textwidth]{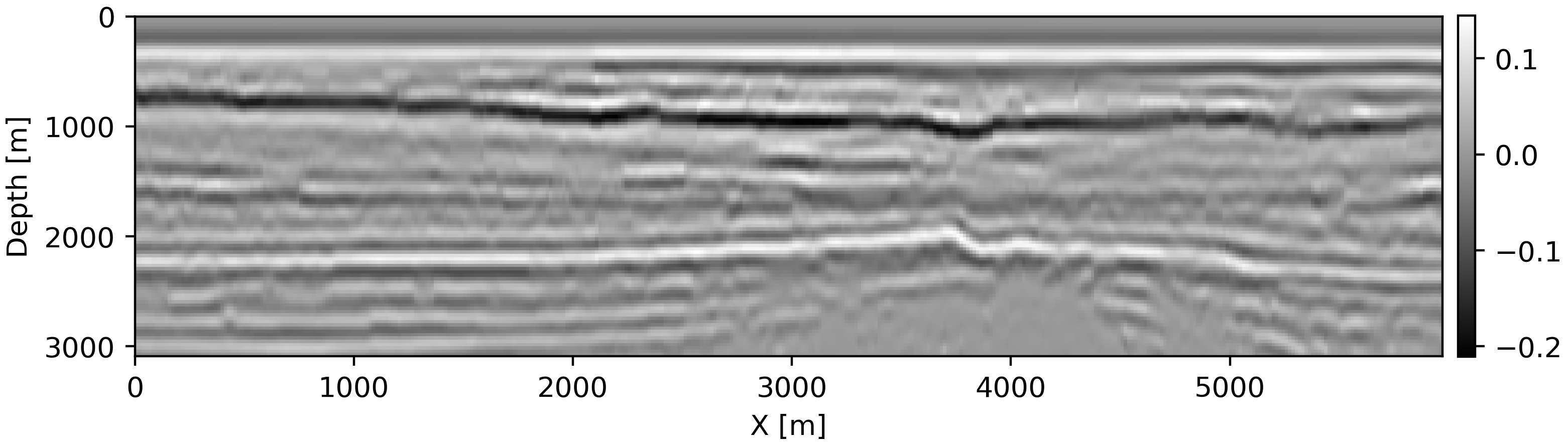}
    \caption{First survey $\mathbf{y}_1$}
    \label{fig:y_1}
\end{subfigure}
\hfill
\begin{subfigure}{0.48\textwidth}
    \centering
    \includegraphics[width=\textwidth]{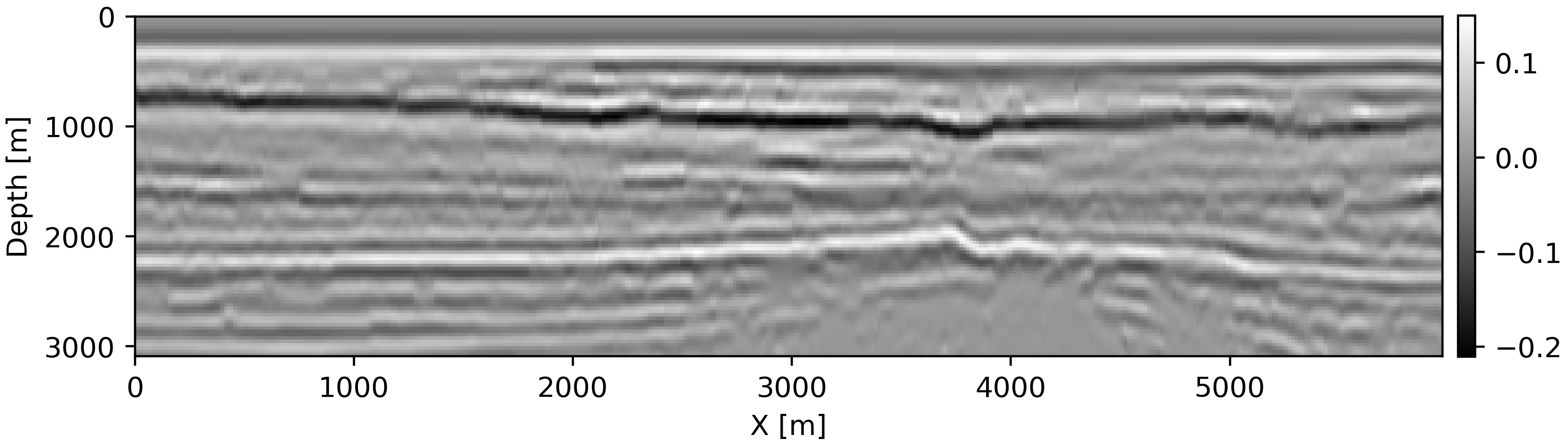}
    \caption{Second survey $\mathbf{y}_2$}
    \label{fig:y_2}
\end{subfigure}
\hfill
\caption{Synthetic case study.  (a) Earth model with CCS injection site and unknown CO$_2$ plume. (b) Simulated post-stack data at first survey. (c) Simulated post-stack data at second survey.  }
\label{fig:case_study}
\end{figure}

Our goal is to recover the $N$ posterior distribution CO$_2$ plume from the $N$ noisy measurements. For simplicity, we illustrate the methodology using $N=2$ time-lapse surveys, though the framework generalizes to $N \geq 2$. 
Mathematically, we aim to solve the two forward models jointly, 
\[
    \mathbf{A}_1\mathbf{x}_1^{\ast} + \epsilon_1 = \mathbf{y}_1 \text{ and } \mathbf{A}_2\mathbf{x}_2^{\ast} + \epsilon_2 = \mathbf{y}_2.
\]
Here, $\mathbf{x}_i^{\ast}$, where $i=1,2$, denotes the true acoustic properties of the subsurface, $\mathbf{y}_i$ denotes the noisy measurement obtained by applying the forward operator $\mathbf{A}_i$ to $\mathbf{x}_i^{\ast}$ with noise $\epsilon_i$. Here we focus on a linear post-stack inversion operator that represents the combination of a convolution operator and a spatial derivative along traces: $\mathbf{A}_i \in \mathbb{R}^{m \times n}$, where $m$ corresponds to the image size and $n$ corresponds to the data size. Examples of the observed data are shown in Figures \ref{fig:y_1} and \ref{fig:y_2}. While this proof of concept uses a linear forward operator, future work will extend the framework to Full-Waveform Inversion, as it is designed to accommodate non-linear operators.

To incorporate uncertainty, we model the solutions probabilistically by using a SGM to decode latent distributions into the posterior estimates $\mathbf{x}_1$ and $\mathbf{x}_2$—i.e., $\mathbf{x}_1 \sim G_\theta (q_{\phi_1}(\mathbf{z}_1)) \text{ and } \mathbf{x}_2 \sim G_\theta (q_{\phi_2}(\mathbf{z}_2))$, where $\mathbf{z}_i \sim \mathcal{N}(0,I)$ is the standard Gaussian distribution, and $q_{\phi_i}$ represents the Gaussian mixture model (GMM) used to encode the different monitors. Consequently, our objective function is defined as follows: 
\begin{align*}
    \min_{\theta, \phi_1, \phi_2}\|\mathbf{A}_1G_\theta(q_{\phi_1}(\mathbf{z}_1)) - \mathbf{y}_1\|_2^2 + \|\mathbf{A}_2G_\theta(q_{\phi_2}(\mathbf{z}_2)) - \mathbf{y}_2\|_2^2 
\end{align*}
which minimizes the difference between the forward-modeled measurements and the observed measurements with respect to the parameters $\theta$ of SGM and $\phi_i$ of the GMMs. After inversion for posterior distributions of $\mathbf{x}_1$ and $\mathbf{x}_2$, we take their difference to make a time-lapse image.

\subsection{Probabilistic Independent Recovery Model}
To evaluate the uplift in performance due to jointly solving $N$ inverse problems versus solving them independently, we also define the Probabilistic Independent Recovery Model (pIRM): 
\begin{align*}
    \min_{\theta_1, \phi_1}\|\mathbf{A}_1G_{\theta_1}(q_{\phi_1}(\mathbf{z}_1)) - \mathbf{y}_1\|_2^2 \text{ and } \min_{\theta_2, \phi_2}\|\mathbf{A}_2G_{\theta_2}(q_{\phi_2}(\mathbf{z}_2)) - \mathbf{y}_2\|_2^2.
\end{align*}
Here a separate generative model is trained for each time-lapse plume, while in pJRM, the generative model is shared thus extracts common features across all time-lapse surveys.

\subsection{Weak formulation} 
When using the pJRM, the computational cost of evaluating the forward operator and its gradient at each iteration is significant. To address this, we adopted a weak formulation introduced by \citeauthor{orozco2024aspireiterativeamortizedposterior}, which decouples forward operator evaluations from network parameter updates in Algorithm~\ref{alg:pjrm}. In this approach, forward operator evaluations and gradient calculations (line 3) occur in an outer loop, while network parameter updates (line 7) take place in an inner loop. This separation reduces the total number of forward operator evaluations required, set to a relatively small value ($\sim200$) compared to the number of network parameter updates needed $\sim100000$. 

\begin{algorithm}
\caption{Probabilistic Joint Recovery Model (pJRM) with Weak Formulation}
\label{alg:pjrm}
\begin{algorithmic}[1]
    \For{$ii = 1 : \text{maxiter}_1$}
        \For{$i = 1 : N$}
            \State $\mathbf{g}_i = \nabla_{\mathbf{x}_i} \left[ \frac{1}{2\sigma^2} \| \mathbf{A}_i(\mathbf{x}_i) - \mathbf{y}_i \|_2^2 + \frac{1}{2\gamma^2} \| \mathbf{x}_i - G_\theta \left( q_{\boldsymbol{\phi}_i} (\mathbf{z}_i) \right) \|_2^2 \right]$
            \State $\mathbf{x}_i = \mathbf{x}_i - \tau \mathbf{g}_i$
        \EndFor
        \For{$iii = 1 : \text{maxiter}_2$}
            \State $\mathcal{L}(\boldsymbol{\phi_j}, \theta) = \sum_{j=1}^N \left[ \frac{1}{2\gamma^2} \| \mathbf{x}_j - G_{\theta} \left( q_{\boldsymbol{\phi_j}} (\mathbf{z}_j)\right) \|_2^2 \right]$
            \State $(\boldsymbol{\phi_j}, \theta)\leftarrow \text{ADAM} (\mathcal{L}(\boldsymbol{\phi_j}, \theta))$
        \EndFor
    \EndFor
\end{algorithmic}
\end{algorithm}

\section{Synthetic case study}
We conducted in-silico validation of our method using velocity models with dimensions of $(3.2 \, \text{Km} \times 5.9 \, \text{Km})$, discretized into a grid of $(398 \times 103)$. As shown in Figure \ref{fig:earth}, we simulated the fluid flow of $\text{CO}_2$ at an injection site using \textit{Jutul}, mimicking a realistic CCS project \citep{jutuldarcy}. The flow of $\text{CO}_2$ alters the acoustic properties of the reservoir, leading to time-lapse differences that can be imaged. To invert these time-lapse images, we employed a post-stack convolutional forward operator using \textit{Pylops} \citep{Ravasi2020PyLops}. We tested the weak formulation of our method using two and six surveys and obtained time-lapse reconstructions by taking the difference between the posterior means. To quantify uncertainty, we plotted the standard deviation of the differences between posterior samples. 
 
\begin{figure}[!htb]
\centering
\begin{subfigure}{0.48\textwidth}
    \centering
    \includegraphics[width=\textwidth]{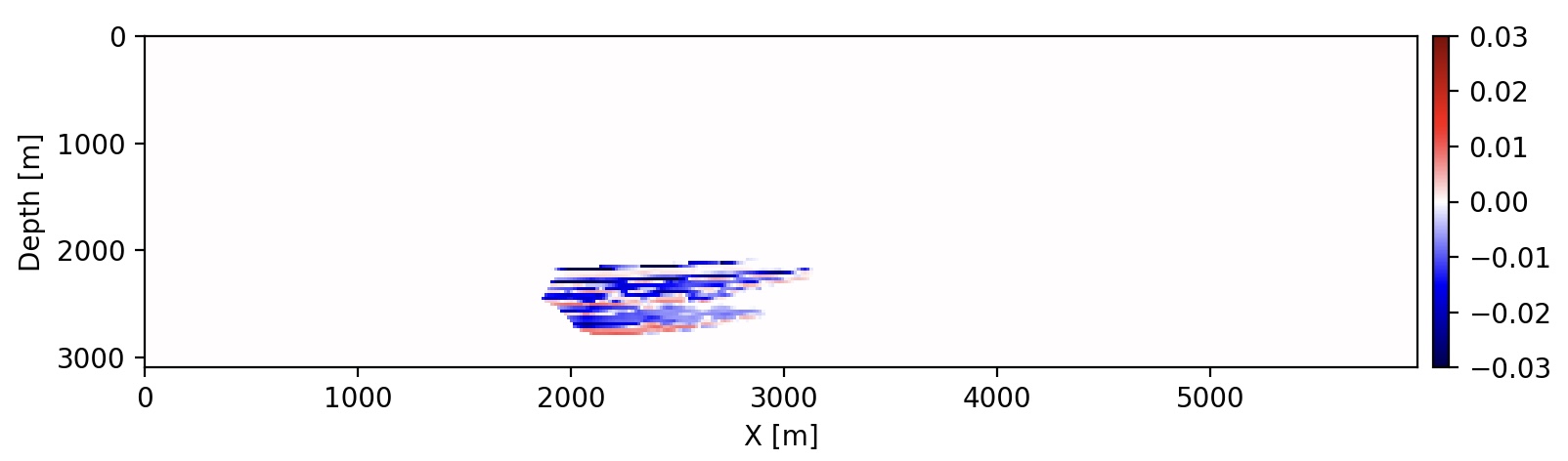}
    \caption{Ground truth time-lapse difference $\mathbf{x}^{*}_2-\mathbf{x}^{*}_1$}
    \label{fig:true_diff}
\end{subfigure}
\hfill
\begin{subfigure}{0.48\textwidth}
    \centering
    \includegraphics[width=\textwidth]{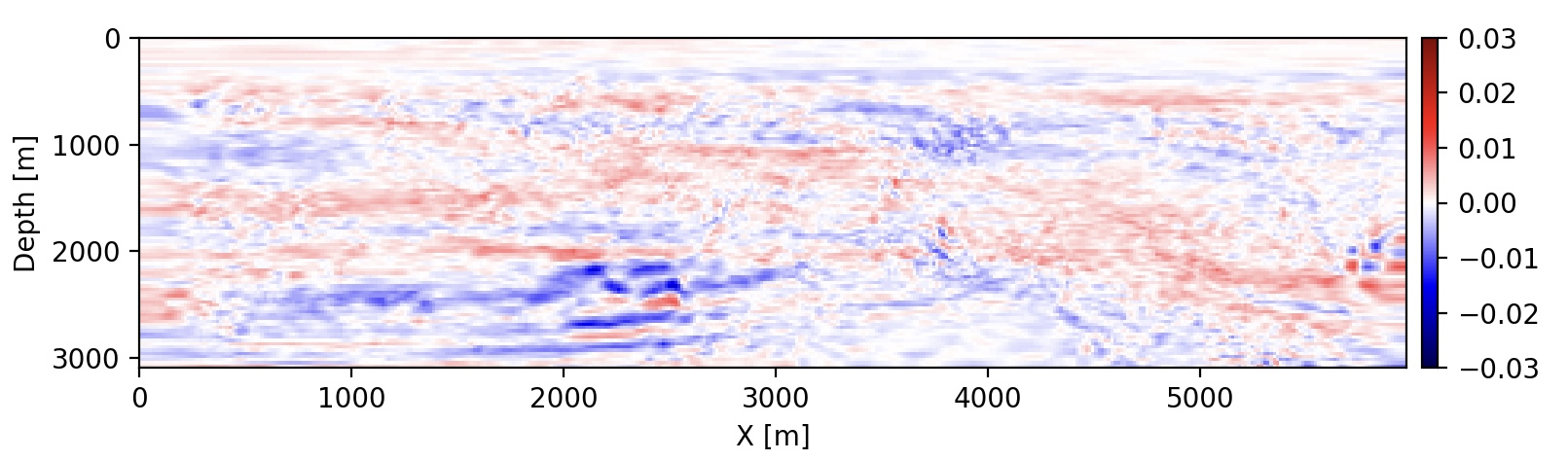}
    \caption{Independent pIRM w/ 2 surveys}
    \label{fig:diff_pIRM}
\end{subfigure}
\begin{subfigure}{0.48\textwidth}
    \centering
    \includegraphics[width=\textwidth]{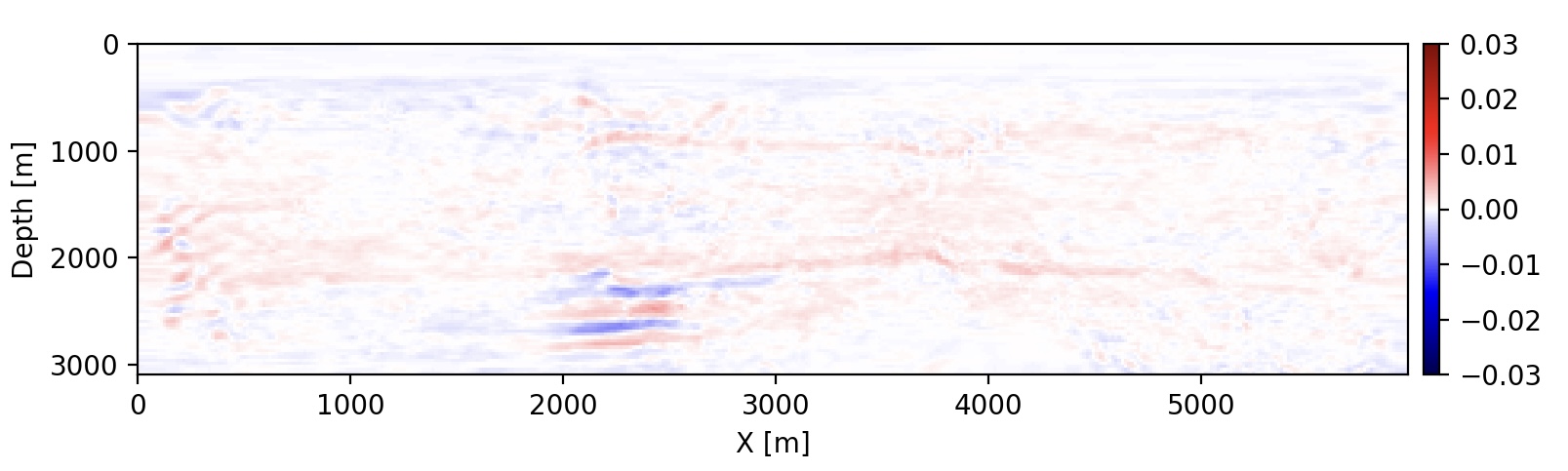}
    \caption{Joint pJRM w/ 2 surveys}
    \label{fig:pjrm_2}
\end{subfigure}
\hfill
\begin{subfigure}{0.48\textwidth}
    \centering
    \includegraphics[width=\textwidth]{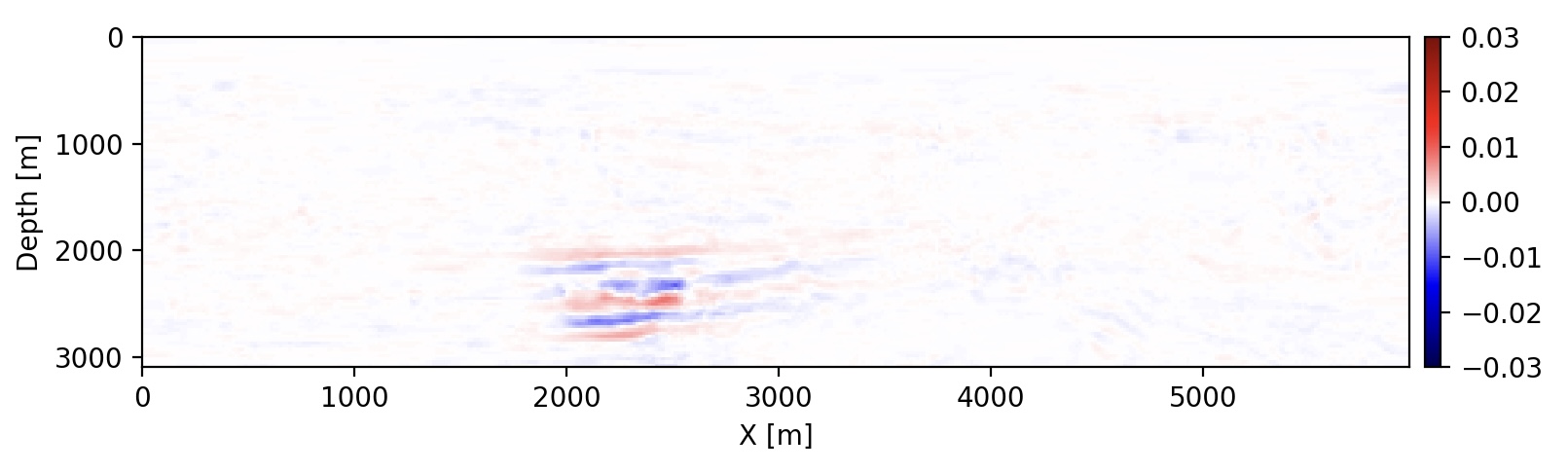}
    \caption{Joint pJRM w/ 6 surveys}
    \label{fig:pjrm_6}
\end{subfigure}

\caption{Comparison of time-lapse Images.  (a) Time-lapse image with Independent \textrm{pIRM}. (b) Ground truth time-lapse difference. (c) Time-lapse image with \textrm{pJRM} and 2 surveys. (d)  Time-lapse image with \textrm{pJRM} and 6 surveys.}
\label{fig:time_lapse_comparison}
\end{figure}

\begin{figure}[!htb]
\centering
\begin{subfigure}{0.48\textwidth}
    \centering
    \includegraphics[width=\textwidth]{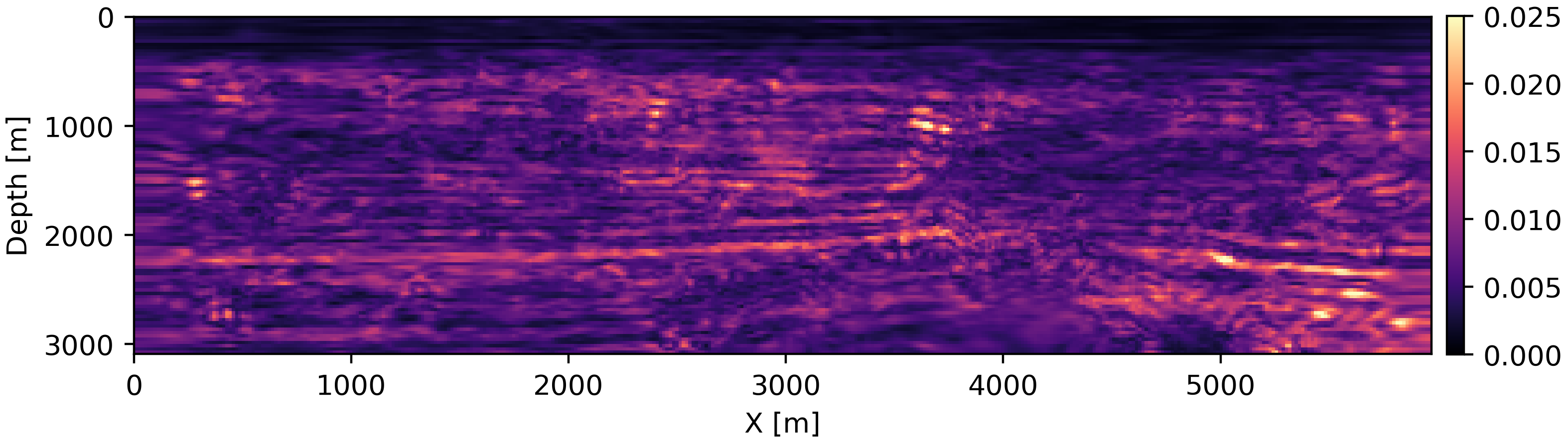}
    \caption{Uncertainty pIRM w/ 2 surveys}
    \label{fig:std1}
\end{subfigure}
\hfill
\begin{subfigure}{0.48\textwidth}
    \centering
    \includegraphics[width=\textwidth]{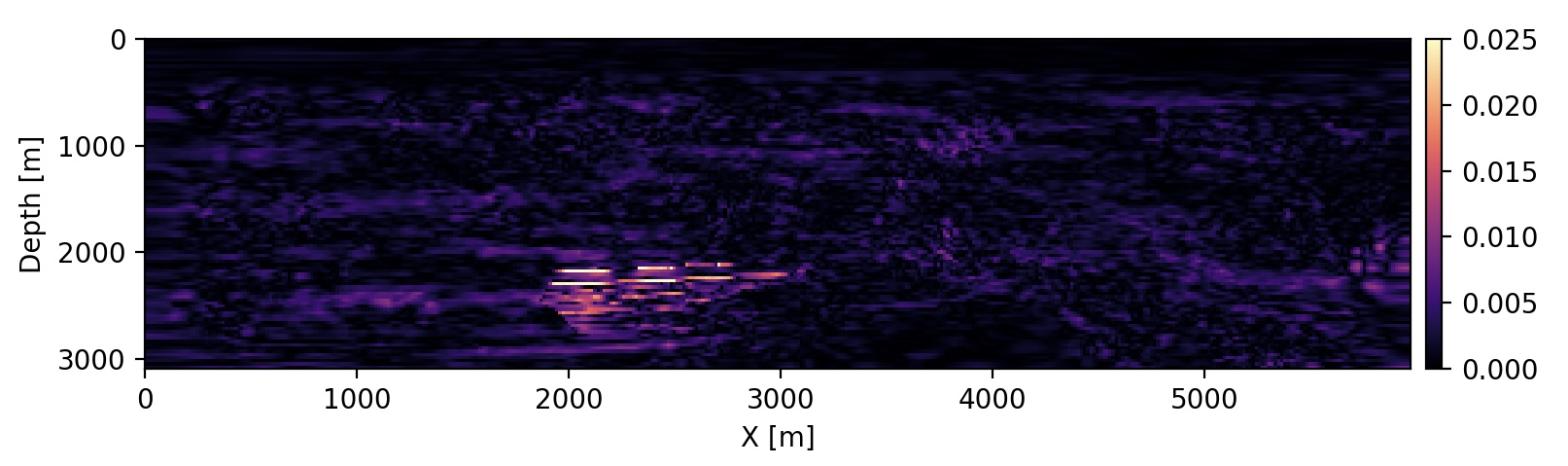}
    \caption{Error pIRM w/ 2 surveys}
    \label{fig:error1}
\end{subfigure}
\begin{subfigure}{0.48\textwidth}
    \centering
    \includegraphics[width=\textwidth]{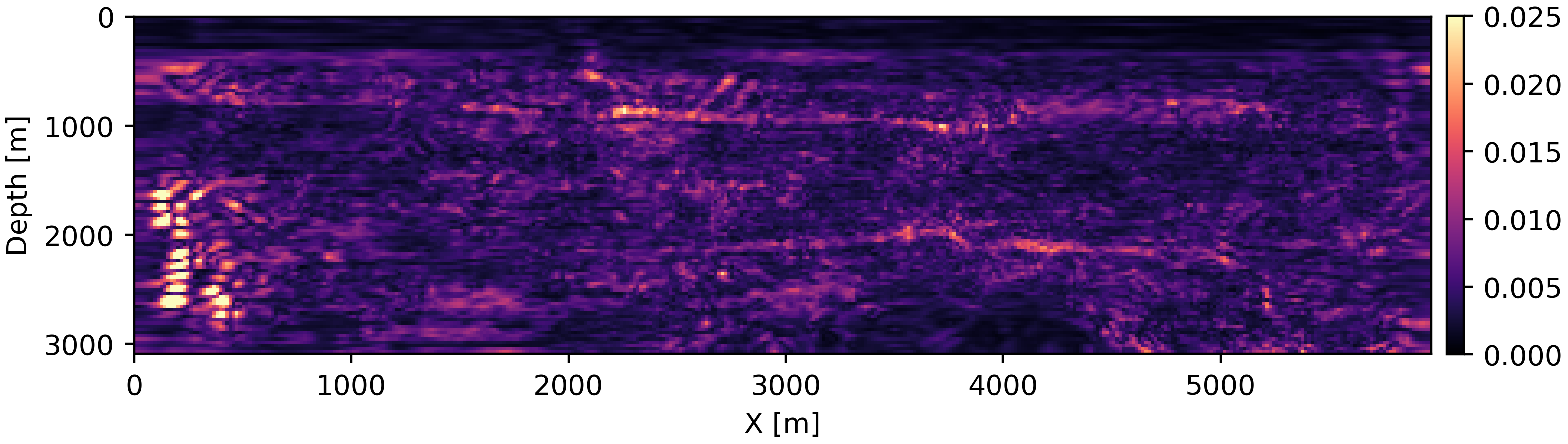}
    \caption{Uncertainty pJRM w/ 2 surveys}
    \label{fig:std2}
\end{subfigure}
\hfill
\begin{subfigure}{0.48\textwidth}
    \centering
    \includegraphics[width=\textwidth]{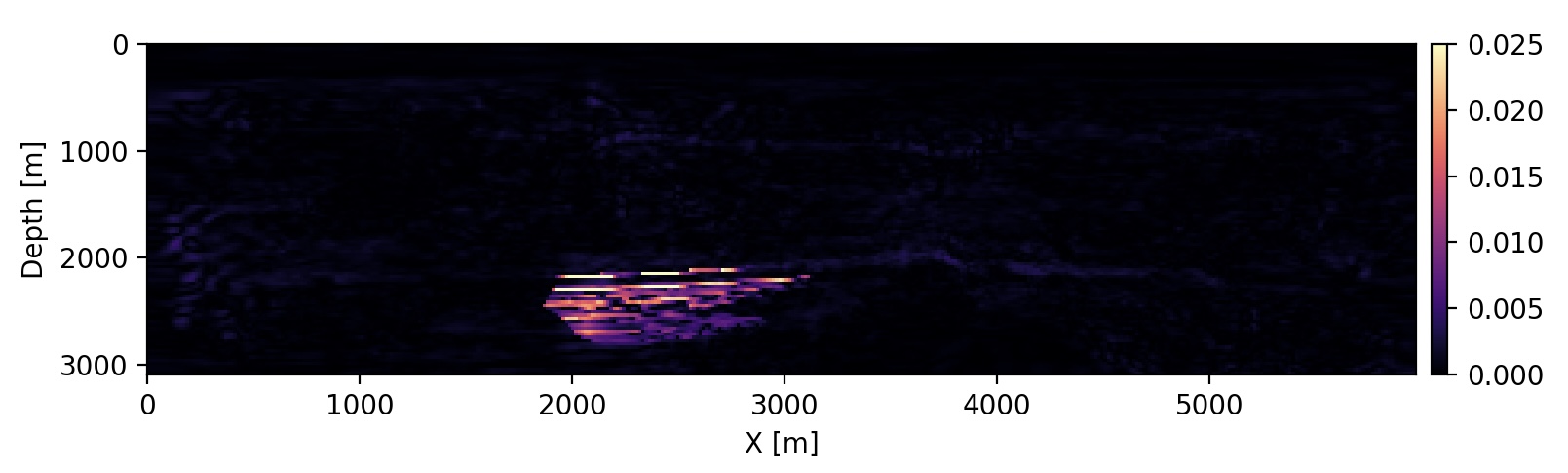}
    \caption{Error pJRM w/ 2 surveys}
    \label{fig:error2}
\end{subfigure}
\begin{subfigure}{0.48\textwidth}
    \centering
    \includegraphics[width=\textwidth]{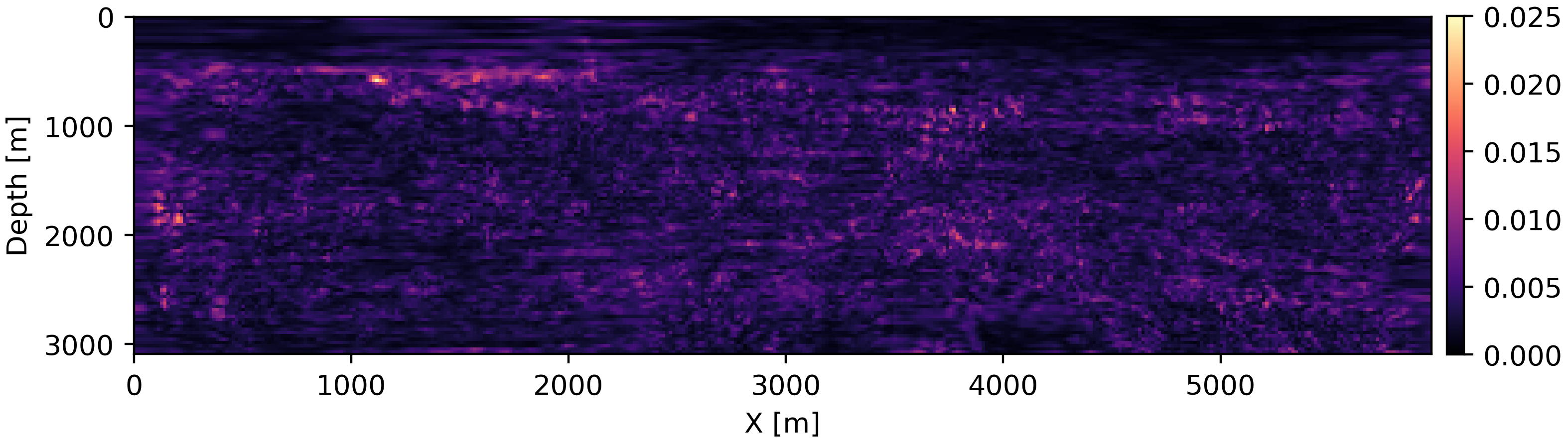}
    \caption{Uncertainty pJRM w/ 6 surveys}
    \label{fig:std3}
\end{subfigure}
\hfill
\begin{subfigure}{0.48\textwidth}
    \centering
    \includegraphics[width=\textwidth]{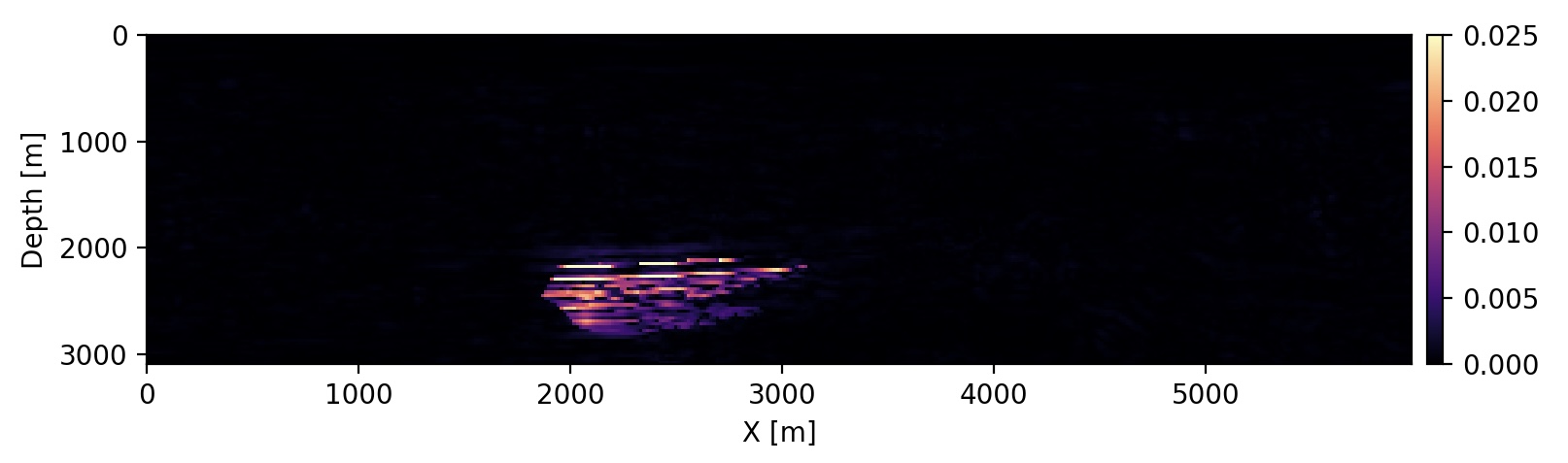}
    \caption{Error pJRM w/ 6 surveys}
    \label{fig:error3}
\end{subfigure}

\caption{Comparison of uncertainty and errors.  (a) Uncertainty of time-lapse image w/ independent method \textrm{pIRM} and 2 surveys. (b) Error of time-lapse image w/ \textrm{pIRM} and 2 surveys.  (c) Uncertainty of time-lapse image w/ joint method \textrm{pJRM} and 2 surveys. (d) Error of time-lapse image w/ \textrm{pJRM} and 2 surveys.  (e) Uncertainty of time-lapse image w/ joint method \textrm{pJRM} and 6 surveys. (f) Error of time-lapse image w/ \textrm{pJRM} and 6 surveys.  }
\label{fig:uq_comparison}
\end{figure}

Juxtaposing Figures \ref{fig:diff_pIRM} and \ref{fig:pjrm_2} demonstrates that pJRM significantly outperforms pIRM, as previously shown in the non-probabilistic case by \citet{gahlot2023IMAGEtsm}. The time-lapse image generated by pJRM retains the overall shape of the CO$_2$ plume with lower uncertainty. In contrast, pIRM fails to capture the plume structure, resulting in high uncertainty both within the plume and across other regions of the reconstruction. Additionally, Figures \ref{fig:pjrm_2} and \ref{fig:pjrm_6} illustrate that increasing the number of surveys improves reconstruction quality, as the generative model benefits from observing the shared components across multiple surveys. Figure \ref{fig:uq_comparison} further highlights a decreasing trend in uncertainties as more surveys are incorporated, validating the expected correlation between uncertainty and error. These results underscore pJRM's effectiveness in representing uncertainty, making it a more reliable tool for monitoring CCS.

\section{Conclusions}
We introduce a novel approach to CCS monitoring that leverages shared structures between surveys to improve reconstruction while providing uncertainty quantification. Additionally, we present a weak formulation of the framework, enabling efficient use of computationally expensive forward operators. Our synthetic experiments highlighted two key findings: first, joint recovery significantly enhances time-lapse signal reconstruction compared to its independently recovered counterpart; second, incorporating additional monitoring surveys further improves performance. Since this approach operates within a probabilistic framework it offers uncertainty analysis for the risk-averse application of CCS monitoring.

\section{Acknowledgments}

This research was carried out with the support of Georgia Research Alliance and partners of the ML4Seismic Center.


\end{document}